\documentclass[letterpaper, 10 pt, conference]{ieeeconf}

\IEEEoverridecommandlockouts

\usepackage{multicol}
\usepackage[bookmarks=true]{hyperref}

\usepackage[tbtags]{amsmath}
\usepackage{amsfonts}
\usepackage{amssymb}
\usepackage{amsthm}
\usepackage{mathrsfs}
\usepackage{xspace}
\usepackage{xparse}
\usepackage{mathtools}
\usepackage{multirow}
\usepackage{graphicx}
\usepackage{xcolor}
\usepackage{booktabs}

\usepackage{url}

\usepackage{tabularx}
\usepackage{algorithm}
\usepackage{algpseudocode}

\title{\LARGE \bf
Zero to Autonomy in Real-Time: Online Adaptation of Dynamics in Unstructured Environments
}

\author{
William~Ward,
Sarah~Etter,
Jesse~Quattrociocchi,
Christian~Ellis,
Adam~J.~Thorpe,
Ufuk~Topcu
\thanks{W. Ward, C. Ellis, A. Thorpe, U. Topcu are with the Oden Institute for Computational Engineering \& Science, University of Texas at Austin. S. Etter and J. Quattrociocchi are with the Department of Computer Science, University of Texas at Austin. J. Quattrociocchi and C. Ellis are with the DEVCOM Army Research Laboratory. 
  Email: {\tt wward@utexas.edu, etter@utexas.edu, christian.ellis@austin.utexas.edu, adam.thorpe@austin.utexas.edu, utopcu@utexas.edu}.
}
\thanks{Corresponding author: W. Ward}
}

\begin{document}

\maketitle

\begin{abstract}
Autonomous robots must go from zero prior knowledge to safe control within seconds to operate in unstructured environments. Abrupt terrain changes, such as a sudden transition to ice, create dynamics shifts that can destabilize planners unless the model adapts in real-time. We present a method for online adaptation that combines function encoders with recursive least squares, treating the function encoder coefficients as latent states updated from streaming odometry. This yields constant-time coefficient estimation without gradient-based inner-loop updates, enabling adaptation from only a few seconds of data. We evaluate our approach on a Van der Pol system to highlight algorithmic behavior, in a Unity simulator for high-fidelity off-road navigation, and on a Clearpath Jackal robot, including on a challenging terrain at a local ice rink. Across these settings, our method improves model accuracy and downstream planning, reducing collisions compared to static and meta-learning baselines.
\end{abstract}

\section{Introduction}

High-speed ground vehicles require dynamics models that evolve as quickly as the terrain itself. When operating near the limits of controllability, even modest prediction errors in ground terrain interaction can lead to instability, skidding, or rollover. This problem is particularly apparent in off-road navigation: transitions such as pavement to loose gravel can change friction properties within seconds, while mixed terrain features introduce variation in the terrain properties that are difficult to accurately predict. Planning frameworks such as model predictive path integral (MPPI) control ~\cite{williams2018mppi} rely on an accurate model of the system dynamics to predict rollouts and select optimal control actions in real-time. As the terrain shifts, the central challenge is that the model must adapt in real-time, at the same rate as the controller, without the luxury of retraining, fine-tuning, or long adaptation windows.
We consider the problem of learning a dynamics model that adapts in real-time to terrain-induced changes in system dynamics.

We employ function encoders, which learn a set of neural basis functions from trajectories across diverse terrains and represent new dynamics as a linear combination of those bases \cite{fe_transfer, ingebrand2024FEnODEs}. The coefficients of the linear combination define the terrain-specific model, allowing rapid adaptation compared to training a new network. Prior work has shown that this representation enables zero-shot transfer to unseen conditions from batches of historical data \cite{ingebrand2024FEnODEs, ward2025FEMPPI}. However, once initialized, the coefficients remain fixed, and the model does not adjust to terrain changes encountered during operation. This leaves the model blind to subsequent terrain changes and limits performance under evolving conditions.

\begin{figure}
    \centering
    \includegraphics[keepaspectratio,width=\linewidth]{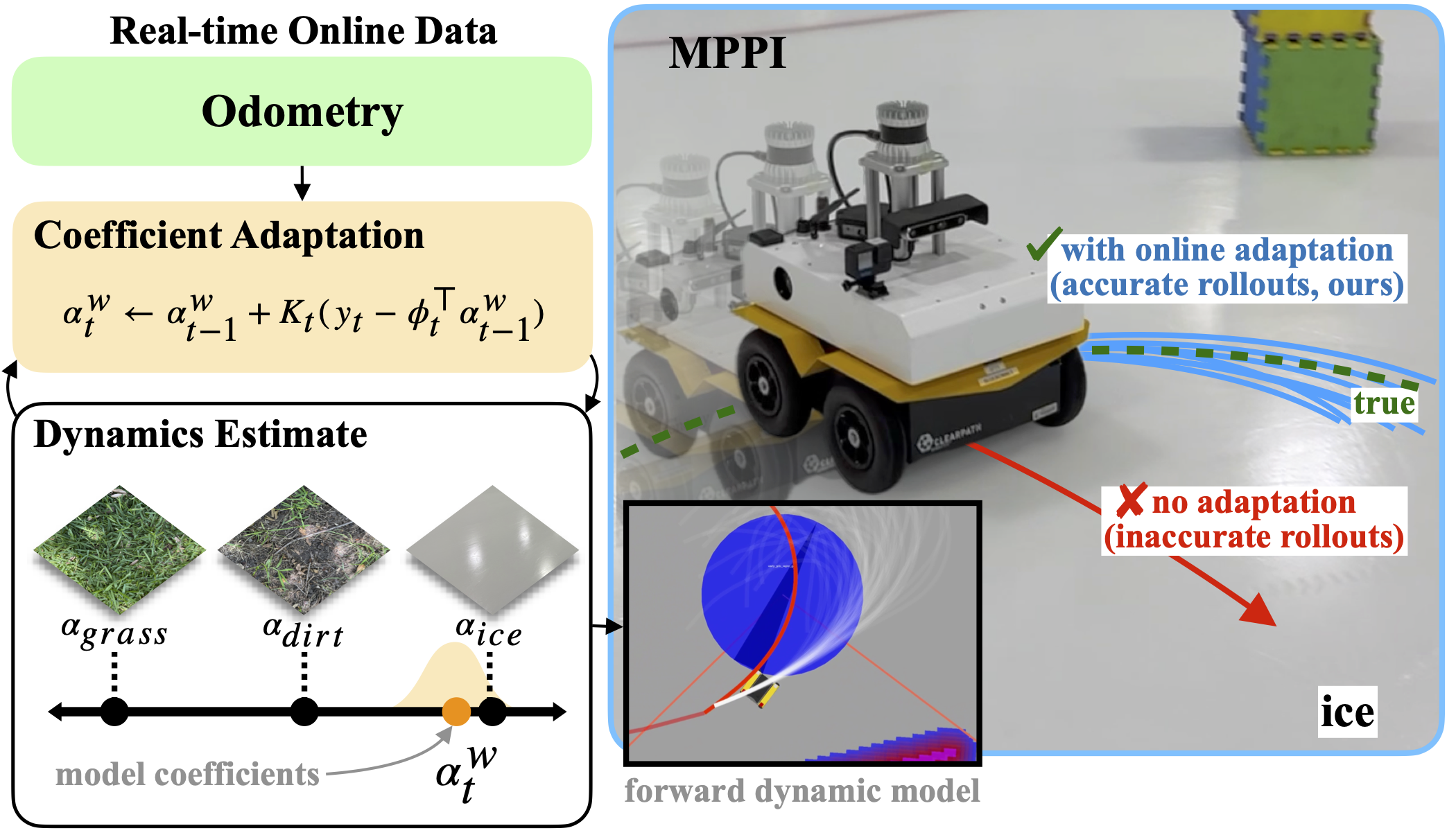}
    \vspace{-0.5cm}
    \caption{Our approach enables rapid adaptation of the dynamics model to a new terrain within seconds. We use recursive least squares to update an online estimate of function encoder coefficients using streaming, real-time data, leading to more accurate rollouts and improved MPPI performance.}
    \label{fig: title figure}
\end{figure}

We propose a method to address this limitation based in function encoders by treating the coefficient vector as a state that is estimated online. We use recursive least squares (RLS) to update the coefficients from streaming data, in the same way that a Kalman filter refines a state estimate (Fig. \ref{fig: title figure}). This makes the dynamics model track terrain changes at the same rate as the controller. The result is a compact, data-driven model that adapts in real-time without retraining. The key distinction from classical adaptive estimation is that adaptation occurs in a learned function space whose basis is trained across environments, rather than in a fixed parametric model. This separates representation learning from online adaptation, enabling real-time updates that generalize across previously observed dynamics. Unlike meta-learning methods such as model-agnostic meta-learning (MAML) ~\cite{finn2017model}, which use gradient-based updates that generally require significantly more data than is feasible for high-speed autonomy, our approach adapts through recursive estimation, not optimization. By constraining learning to the coefficient space of function encoders, recursive estimation yields a model adapted to the observed terrain in real-time.

% Paragraph 4: Contributions
Our main contribution is a method for obtaining an adaptive, learned model of system dynamics that can be updated through RLS at the same rate as the controller. We integrate this model with MPPI and evaluate it in both simulation and on hardware. In simulation, we use a Unity-based ROS simulator. On hardware, we test our approach on a Clearpath Jackal. In both settings, we evaluate performance on adversarial ice environments that push the limits of controllability. While the Jackal operates at relatively modest speeds, our experiments reveal a qualitative difference in the planning capability of MPPI when using our adaptive model, highlighting the importance of model accuracy. Our results show that learned dynamics models can adapt at the timescales needed for control by combining function encoders, neural ODE basis functions, and RLS.

\subsection{Related Work}

\subsubsection*{Terrain parameter estimation}
Existing work estimates terrain-specific parameters such as slip, sinkage, or deformability directly from sensor data \cite{pentzer2014online, li2018multi, padmanabhan2018estimation, espinoza2019vehicle, reina2020terrain, dallas2020online}. A separate line of work bypasses terrain modeling entirely by learning traversability classifiers \cite{Meng-RSS-23}, which predict whether terrain is passable but do not capture the underlying dynamics needed for control. These approaches are data-efficient and interpretable, but rely on fixed (often analytic) underlying dynamics models. When true terrain dynamics deviate from this assumed structure, predictions degrade and planning reliability suffers.

\subsubsection*{Learning-based estimation and prediction}
Data-driven dynamics models have succeeded in domains such as robotic manipulation \cite{byravan2017se3, gillespie2018learning}, structured autonomous driving \cite{spielberg2019neural, nie2022deep}, drone navigation~\cite{pmlr-v229-djeumou23a}, and dynamic maneuvers on controlled surfaces \cite{djeumou2023drift}. These approaches replace analytical models with neural network-based models, but they struggle to generalize when deployed on terrains absent from training data. Assumptions of consistent ground conditions often break down in unstructured off-road environments, where dynamics vary rapidly and unpredictably.

Learning-based methods leveraging transformer architectures can generalize across environments by encoding historical state–action data into a latent context vector \cite{wang2024payattn, xiao2025anycar}.
Similarly, policy adaptation methods such as Rapid Motor Adaptation (RMA) \cite{kumar2021rma} predict latent embeddings that capture terrain properties and enable adaptation without gradient updates.
Conditioning the policy on latent representations using trajectory data enables adaptation and improves predictive accuracy. However, these methods typically require large-scale training data from simulation and rely on privileged information about the simulation environment (e.g.\ surface friction and center-of-mass) that may be inaccessible during training. In contrast to the implicit, black-box representation used in RMA, our approach performs explicit, closed-form updates in a learned function space, maintaining a structured and interpretable coefficient representation that enables direct coefficient estimation rather than relying on a neural trajectory encoder.

% \subsubsection*{Function encoders}
Function encoders provide a representation that combines the flexibility of learned models with the efficiency of parameter estimation. By representing dynamics as a linear combination of learned basis functions \cite{ingebrand2024pmlr}, they enable zero-shot transfer from limited data. Prior work demonstrated their potential for terrain-aware planning in simulation \cite{ward2025FEMPPI}, but adaptation occurred only at initialization, leaving the model static during operation.

\subsubsection*{Transfer learning and meta-learning}
A common strategy for handling new tasks or terrains is to fine-tune pre-trained models on new data. Meta-learning approaches extend this idea by preparing models for rapid adaptation with limited samples. Methods such as MAML \cite{finn2017model} and HyperDynamics \cite{xian2021hyperdynamics} have shown promise in non-stationary tasks, and relate closely to lifelong learning approaches \cite{liu2020learning, liu2021lifelong}. 
However, these methods ultimately depend on gradient-based updates, limiting their ability to provide real-time online adaptation for systems requiring high-speed control.
For example, MAML adapts by applying inner-loop gradient updates on new samples via backpropagation and several gradient steps per adaptation, but its performance depends directly on the number of samples available for the update. In contrast, function encoders reduce adaptation to updating a small coefficient vector, while RLS provides an analytic, fixed per-sample cost update rule that is independent of data length.

\subsubsection*{Online learning and adaptive estimation}
Classical online adaptation methods such as RLS and Kalman filters refine parameter estimates efficiently as new data arrives, but require strong structural assumptions. Neural approaches such as online stochastic gradient descent offer more flexibility, but demand substantial compute and are difficult to run at control frequencies. Hybrid methods, including kernel-based adaptation \cite{10885815} and Gaussian process (GP) regression \cite{Wang2005GP}, balance data efficiency with generalization but remain computationally heavy at scale. Our work combines these perspectives: we constrain estimation to the low-dimensional coefficient space of function encoders, and adapt this coefficient vector online using RLS. This enables real-time adaptation without gradient updates or retraining.

\section{Preliminaries \& Problem Formulation}

We seek to learn a dynamics model that can adapt in real-time to unfamiliar and rapidly changing dynamics using streamed real-time data. This is especially challenging in off-road navigation, where a robot may rapidly encounter diverse terrains with drastically different dynamics. We focus on MPPI control, a sampling-based method that relies on a dynamics model to simulate candidate control sequences and optimize the controlled trajectory. Below, we formalize the MPPI problem, but refer the reader to \cite{williams2017information} for further details.

\subsection{System Dynamics and Control}

We consider a discrete-time system influenced by an unknown world state $w \in \mathcal{W}$ that captures terrain effects, with dynamics given by
\begin{equation}
    \label{eqn: system dynamics}
    x_{t+1} = F^{w}(x_{t}, v_{t}), 
\end{equation}
where $x_{t} \in \mathbb{R}^{n}$ is the state of the system and $v_{t} \in \mathbb{R}^{m}$ is the control input at time $t$. 
Since the world state $w$ is unobserved, the true dynamics $F^{w}$ are unknown. The control input $v_t$ applied to the system is corrupted by Gaussian noise with covariance $\Sigma$, such that $v_{t} \sim \mathcal{N}(u_{t}, \Sigma)$. Therefore, $v_t$ is not directly controllable, and we instead optimize a mean control sequence $U = \lbrace u_{0}, u_{1}, \ldots, u_{T-1} \rbrace$ over a horizon $T\in\mathbb{N}$.

We define a terminal cost $\psi(x_{T})$ and a running cost $\mathcal{L}(x_{t}, u_{t}) = c(x_{t}) + \frac{1}{2} u_{t}^{\top} R u_{t}$. The finite-horizon optimal control problem is
\begin{equation}
\label{eq: opt control problem}
\min_{U \in \mathcal{U}} \mathbb{E}\Bigl[\psi(x_{T}) + \sum_{t=0}^{T-1}\mathcal{L}(x_{t}, u_{t})\Bigr],
\end{equation}
where the expectation is taken over sampled control sequences $V = {v_{0}, \ldots, v_{T-1}}$, subject to the dynamics in \eqref{eqn: system dynamics}.

MPPI control \cite{williams2017information} solves \eqref{eq: opt control problem} by simulating rollouts under a candidate dynamics model $\hat{F}^{w}$. For $r$ sampled sequences $V$, each trajectory $\lbrace x_{1},\ldots,x_{T} \rbrace$ evolves as $x_{t+1} = \hat{F}^{w}(x_{t}, v_{t})$, with rollout cost
\begin{equation}
\label{eqn: state cost}
S(V, x_{0}) = \psi(x_{T}) + \sum_{t=0}^{T-1} c(x_{t}).
\end{equation}
The control update is obtained as a weighted average of sequences, followed by a smoothing step. Importantly, MPPI is model-agnostic, meaning it requires only the ability to forward-simulate trajectories. However, performance depends critically on the accuracy of $\hat{F}^{w}$.

\subsection{Problem Statement}

In off-road navigation, mismatch between the learned model $\hat{F}^{w}$ and the true dynamics $F^{w}$ arises frequently. Terrain properties such as friction, slip, and roughness can change within seconds, and the unobserved world state $w$ prevents direct measurement of these effects. Without an adaptation mechanism, static models trained offline are limited to the distribution of training environments and therefore cannot handle such variability. 

We consider the problem of learning a dynamics model $\hat{F}^{w}$ that adapts in real-time to changes in the unobserved world state. The model must update continuously from streaming data, match the timescale of MPPI, and remain computationally efficient enough for deployment on hardware. 

We assume access to a training dataset, consisting of historical trajectories collected during robot operation on diverse terrains. From the trajectory data, we construct a set of datasets $\lbrace D^{w_1}, \ldots, D^{w_\ell} \rbrace$. Each dataset $D^{w_i}$ consists of states and actions collected from a single dynamics function $F^{w_i}$, $D^{w_i} = \left\lbrace (x_j, v_j, x_{j+1}) \right\rbrace_{j=0}^{m}$,
where $x_{j+1} = F^{w_i}(x_j, v_j)$. 
At runtime, the system receives new samples sequentially from online odometry measurements to refine an estimate $\hat{F}^{w}$.

\section{Methods}

\subsection{Learning an Adaptable Model Using Function Encoders}

We represent system dynamics using function encoders \cite{ingebrand2024FEnODEs}, which provide a compact basis-function representation that supports adaptation through coefficient updates.

Let $\mathcal{F} = \lbrace f^{w} \mid w \in \mathcal{W} \rbrace$ be the space of all dynamics parameterized by the world state $w \in \mathcal{W}$. 
The evolution of the system follows 
\begin{equation}
    x_{t+1} = F^{w}(x_{t}, v_{t}) = x_{t} + \int_{t}^{t+1} f^{w}(x(\tau), v_{t}) d \tau,
\end{equation}
where $f^{w}$ is the underlying vector field, and we integrate $f^{w}$ holding $v_{t}$ constant over the interval $[t, t+1]$. 

A function encoder learns a set of neural ODE basis functions $\lbrace g_{1}, \ldots, g_{k} \rbrace$ that span a subspace $\hat{\mathcal{F}} = \mathrm{span}\lbrace g_{1}, \ldots, g_{k} \rbrace$ supported by the data. 
The dynamics $f^{w}$ are approximated as $\hat{f}^{w} \in \hat{\mathcal{F}}$. Any function $\hat{f}^{w}$ takes the form of a linear combination of the learned basis functions,
\begin{equation}
    \label{eqn: function encoder dynamics estimate}
    \hat{f}^{w}(x, v) = \sum_{j=1}^{k} \alpha_{j}^{w} g_{j}(x, v; \theta_{j}),
\end{equation}
where $w$ represents the world state, $\alpha^{w} \in \mathbb{R}^{k}$ are the real coefficients corresponding to the dynamics in the learned space, and $\theta_{j}$ are the parameters of the neural ODE $g_{j}$. 
Intuitively, the basis functions capture general patterns of the dynamics on different terrains, while the coefficients specialize these patterns to specific terrains.

We presume access to a set of datasets $\lbrace D^{w_{1}}, \ldots, D^{w_{\ell}} \rbrace$ consisting of observed trajectories (transitions) across different terrains during training. 
For each dataset $D^{w_{i}}$, we compute coefficients $\alpha^{w_{i}}$ via least squares, estimate the dynamics as in \eqref{eqn: function encoder dynamics estimate}, and compute the prediction or reconstruction loss as the mean squared error of the change in state. We minimize the average reconstruction error across all terrains via gradient descent. After training, the basis functions remain fixed.
For more training details, we refer the reader to \cite{ingebrand2024FEnODEs, fe_transfer}. 

During training, the neural ODE basis functions learn to span the subspace supported by the data and capture the different terrain-induced dynamics functions. Once the basis is trained, the coefficients $\alpha^{w}$ provide a compact representation of a particular terrain or environment. While function encoders enable zero-shot transfer through offline coefficient estimation, they remain static during deployment. The proposed RLS updates convert this representation into a continuously adapting model, allowing the coefficients to track evolving dynamics at control rates.

Prior work estimates the coefficients using a least-squares solve. The coefficients $\alpha^{w}$ can be computed in closed-form via the normal equation as $(G + \lambda I) \alpha = F$, where $G_{ij} = \langle g_{i}, g_{j} \rangle$ and $F_{i} = \langle g_{i}, f^{w} \rangle$, computed using Monte Carlo integration using a small amount of input-output data of the form $(x_{t}, v_{t})$ and $x_{t+1} - x_{t}$ \cite{ward2025FEMPPI, ingebrand2024FEnODEs}. 
Though this procedure works well to parameterize the dynamics on different terrains, it creates a computational bottleneck in online settings. 
Computing the coefficients requires a batch of data, computing a $k \times k$ matrix, and then computing the matrix inverse. This process is only possible once the robot has driven on a terrain long enough to gather sufficient data. Additionally, the coefficients correspond to only a single terrain, and remain fixed until we perform another batch solve.
In off-road driving, terrain transitions rarely occur with clear boundaries. Conditions can shift rapidly, making it impractical to pause and recompute coefficients. This reliance on lump-sum data and repeated batch solves limits the function encoder model's ability to adapt in real-time.

\subsection{Computing the Coefficients Using RLS}

To enable real-time adaptation, we treat the coefficients $\alpha^w$ as a latent state and estimate them using RLS. Intuitively, the updates follow the same structure as Kalman filtering: $\alpha^w$ plays the role of the state, the covariance matrix $P_t$ tracks uncertainty in this estimate, and the Kalman gain $K_t$ determines how strongly each new measurement influences the update. 

At each time instant, the robot collects an input-output measurement of the form $(x_{t}, v_{t}, x_{t+1})$ from onboard odometry.
Let $y_{t} = x_{t+1} - x_{t}$ and $\Phi_{t} \in \mathbb{R}^{n \times k}$ with columns $g_{k}(x_{t}, v_{t})$.
The dynamics model takes the form, $y_{t} = \Phi_{t} \alpha^{w} + \epsilon$, where $\epsilon \sim \mathcal{N}(0, Q)$ denotes measurement noise. Our goal is to update the coefficients as sequential data arrives.

Specifically, we initialize the coefficient estimate at $t=0$ as $\alpha_{0}^{w} = \boldsymbol{0}$, which represents zero prior knowledge, and the covariance $P_{0} = \lambda^{-1} I$, where $\lambda > 0$ is the regularization parameter. 
Let $\gamma \in (0, 1]$ be a forgetting factor to reduce the value of older measurements in our estimate, and let $P_{t | t-1} = \frac{1}{\gamma} P_{t-1}$. 
At each time step, we update the coefficients using the following update equation, 
\begin{equation}
    \alpha_t^{w} = \alpha_{t-1}^{w} + K_t (y_t - \Phi_t \alpha_{t-1}^{w}).
\end{equation}
Let $S_{t} = \Phi_{t} P_{t | t-1} \Phi_{t}^{\top} + R$. The gain $K_t = P_{t | t-1} \Phi_t^{\top} S_{t}^{-1}$ adjusts how much the coefficients shift in response to the prediction error $y_t - \Phi_t \alpha_{t-1}^{w}$, and the covariance $P_t = P_{t | t-1} - K_t \Phi_t P_{t | t-1}$ contracts over time, meaning updates become smaller as confidence grows. 

We summarize the RLS procedure for computing the coefficient updates in Algorithm \ref{algo: RLS coefficient update}. 

\begin{algorithm}[!ht]
\begin{algorithmic}[1]
\State \textbf{Initialize:} $\alpha_0 = 0$, $P_0 = \lambda^{-1} I$, $\gamma \in (0, 1]$, noise $Q$
\For{each time step $t = 1, 2, \dots$}
    \State $P_{t | t-1} = \frac{1}{\gamma} P_{t-1}$
    \State $S_{t} \gets \Phi_{t} P_{t | t-1} \Phi_{t}^{\top} + Q$
    \State Compute gain: $K_t \gets P_{t | t-1} \Phi_t^{\top} S_{t}^{-1}$
    \State Update coefficients: $\alpha_t^{w} \gets \alpha_{t-1}^{w} + K_t (y_t - \Phi_t \alpha_{t-1}^{w})$
    \State 
    Update covariance: $P_t \gets P_{t | t-1} - K_t \Phi_t P_{t | t-1}$
\EndFor
\end{algorithmic}
\caption{RLS Coefficient Updates}
\label{algo: RLS coefficient update}
\end{algorithm}

Na{\"i}vely, we could recompute the normal equations at every time step. However, recomputing the least-squares solution at every step would require inverting a $k \times k$ matrix that costs $\mathcal{O}(k^3)$, which is generally too expensive for real-time. RLS avoids this by updating both the coefficient estimate and its covariance directly.
By reducing adaptation to a closed-form $\mathcal{O}(k^2)$ update of the coefficient vector, RLS enables function encoders to evolve continuously with incoming data.

\subsection{Real-Time Adaptation in Model Predictive Control}

We integrate the RLS procedure directly into the MPPI control loop. At each time step, the robot collects a new odometry measurement $(x_{t}, v_{t}, x_{t+1})$. We use this observed transition to update the coefficients $\alpha_{t}^{w}$ via the RLS equations as in Algorithm \ref{algo: RLS coefficient update}. The coefficients $\alpha_{t}^{w}$ define the current terrain-aware dynamics $\hat{f}^{w}$. 

At each control step, MPPI uses the current model $\hat{f}^{w}$ to simulate candidate rollouts. For each sampled control sequence, we compute the state trajectories, compute trajectory costs, and update the nominal control sequence as normal. 
Because coefficients update immediately as new data arrives, the model remains aligned with actual vehicle dynamics on the current terrain, even when the terrain changes mid-operation.

This coupling between recursive coefficient estimation and sampling-based planning ensures that the dynamics model evolves at the same timescale as control. In effect, the RLS updates serve as a lightweight, Kalman filter-like estimator for the latent terrain coefficients, while MPPI leverages the updated model for trajectory optimization.

\section{Results}
We design three experiments that progressively increase in complexity to evaluate real-time adaptation using function encoders with RLS (FE-RLS).
First, a Van der Pol oscillator provides a controlled setting to isolate the adaptation mechanism and compare against gradient-based meta-learning.
Second, a Unity-based ROS simulator provides a high-fidelity environment for testing quantitative and qualitative performance under controlled terrain changes.
Finally, hardware experiments on a Clearpath Jackal validate our approach under real-world conditions, including a challenging ice rink setting.

We compare against three representative baselines.
The first baseline is a function encoder (FE) with access to unrealistic batch data sampled uniformly from the test terrain dataset, which computes the coefficients offline using least squares.
It serves as a useful reference but not a firm lower bound, since our online method (FE-RLS) can leverage additional data at runtime to refine coefficients.
The second is a neural ODE \cite{chen2018neuralODEs} trained on the same data as the function encoder. The neural ODE remains fixed without any online adaptation, illustrating the performance of a standard learned dynamics model that remains static after training. 
The third is MAML \cite{finn2017model}, which represents the dominant approach for fast adaptation.
The MAML-adapted neural ODE trains a single neural network so that its parameters are easily adaptable to new environments.

\subsection{An Illustrative Example on a Van der Pol Oscillator}

\begin{figure}
    \centering
    \includegraphics[width=\linewidth]{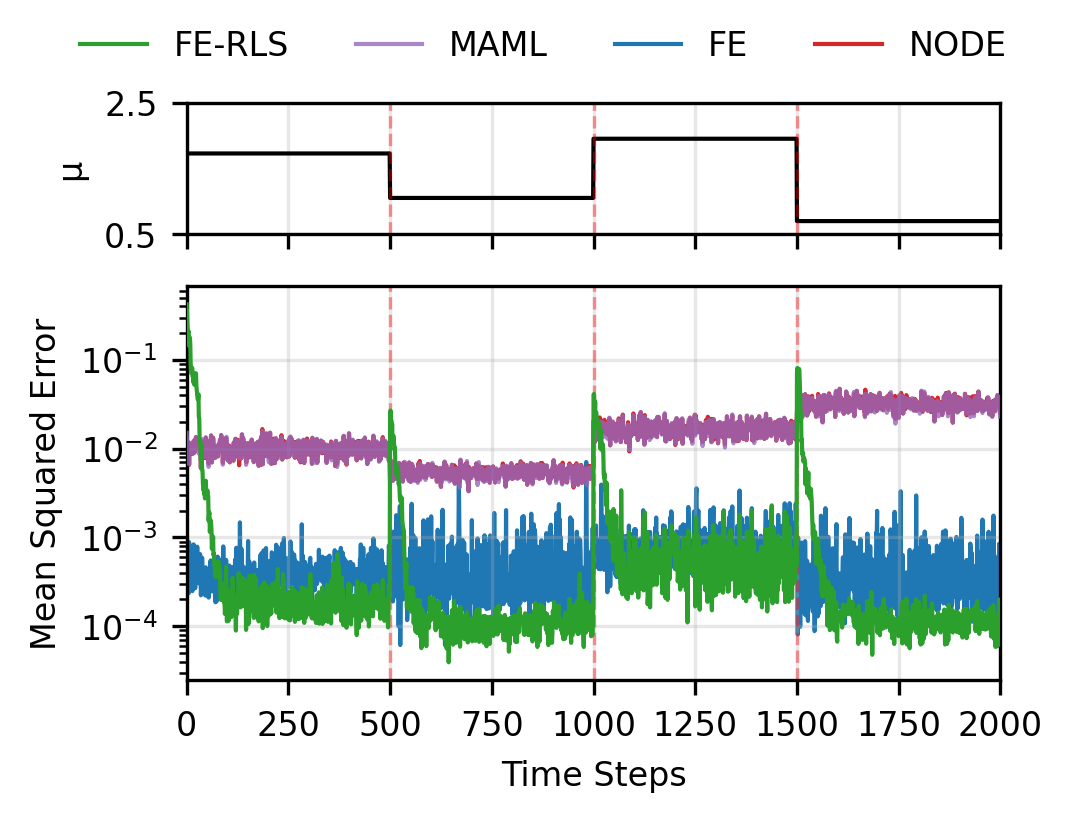}
    \vspace{-0.5cm}
    \caption{Adaptation performance on the Van der Pol system as the $\mu$ parameter changes. FE-RLS rapidly converges to an accurate dynamics representation within $\approx 5$ seconds after $\mu$ changes, while MAML (5 inner steps, 100 data point buffer) does not meaningfully adapt under single-step updates. NODE and MAML have near-identical performance.}
    \label{fig: van der pol comparison}
\end{figure}

We first evaluate the method on the Van der Pol oscillator to illustrate the adaptation algorithm.
The system dynamics are given by 
$\dot{x}_1 = x_2$, $\dot{x}_2 = \mu (1 - x_1^2) x_2 - x_1$,
where $\mu$ is an unobserved parameter (similar to unobserved terrain parameters).
Using the training methodology from~\cite{ingebrand2024FEnODEs}, we train a function encoder on trajectories generated from different $\mu$ values, then evaluate online adaptation on a system with a new, unseen $\mu$.

At test time, FE-RLS begins with coefficients set to zero ($\alpha^{w} = \boldsymbol{0}$) and updates the coefficients recursively from streaming data using Algorithm \ref{algo: RLS coefficient update} to approximate the dynamics for the Van der Pol system as $\mu$ changes.
For MAML, we vary the number of samples available for inner-loop adaptation to test how quickly it can recover the new dynamics. 
We note that MAML generally requires more online data to adapt and is not designed to adapt using single measurements at every time step. We used 5 inner loop gradient updates with varying-length buffers of data (1, 5, 50, and 100 samples). 
In our experiments, MAML did not effectively adapt to the streaming data, with only a 4\% improvement over the static neural ODE. 

Figure \ref{fig: van der pol comparison} shows that FE-RLS rapidly identifies the correct coefficients for the unseen $\mu$ using $\approx 50$ data points, corresponding to only 5 seconds of data collection. In contrast, MAML does not adapt effectively when restricted to single-step updates and online data. MAML's error closely matches that of the static NODE baseline (the two signals overlap in the figure), indicating minimal adaptation from its initial parameters. The batch function encoder highlights that recursive estimation achieves comparable accuracy to batch coefficient calculations while operating fully online, even achieving lower mean squared error since it effectively uses more data online. 

\subsection{High Fidelity Simulation Results}
    \label{sec: sim results}

\begin{figure}
    \centering
    \includegraphics{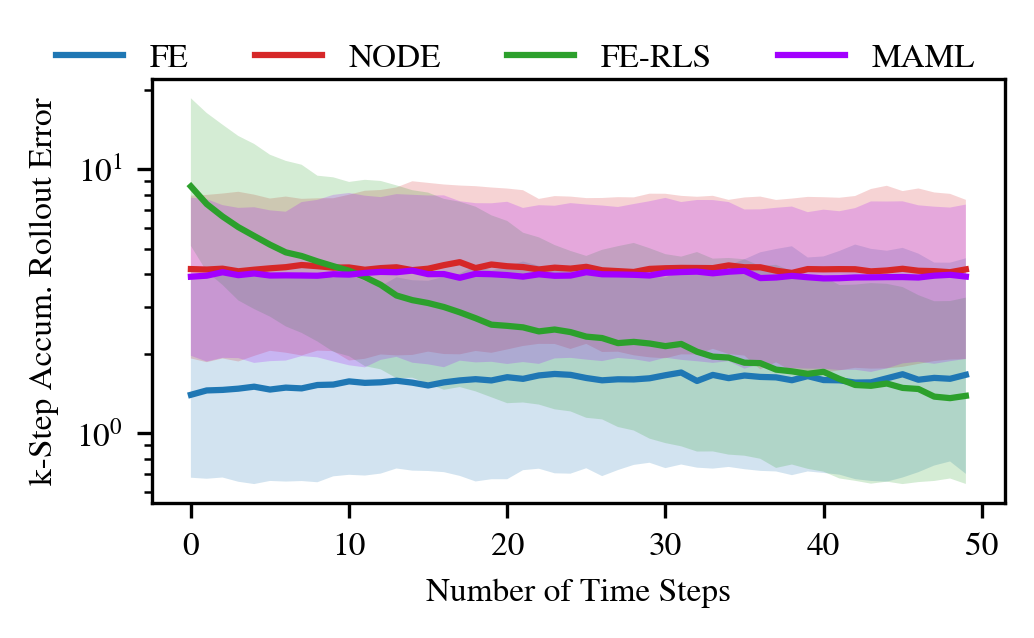}
    \vspace{-0.5cm}
    \caption{At each time step, we compute the accumulated prediction error over the next $k$ time steps, simulating the accuracy of MPPI rollouts at the current time step. FE-RLS adapts to an unknown icy terrain within 5 seconds (1 time step $\approx 0.1$ seconds), starting from zero prior knowledge, while MAML fails to meaningfully adapt. The FE baseline shows the model error with coefficients computed with privileged information, using 100 data points from the terrain collected offline.
    }
    \label{fig: sim k-step}
\end{figure}

\begin{figure}
    \centering
    \includegraphics[keepaspectratio,width=\columnwidth]{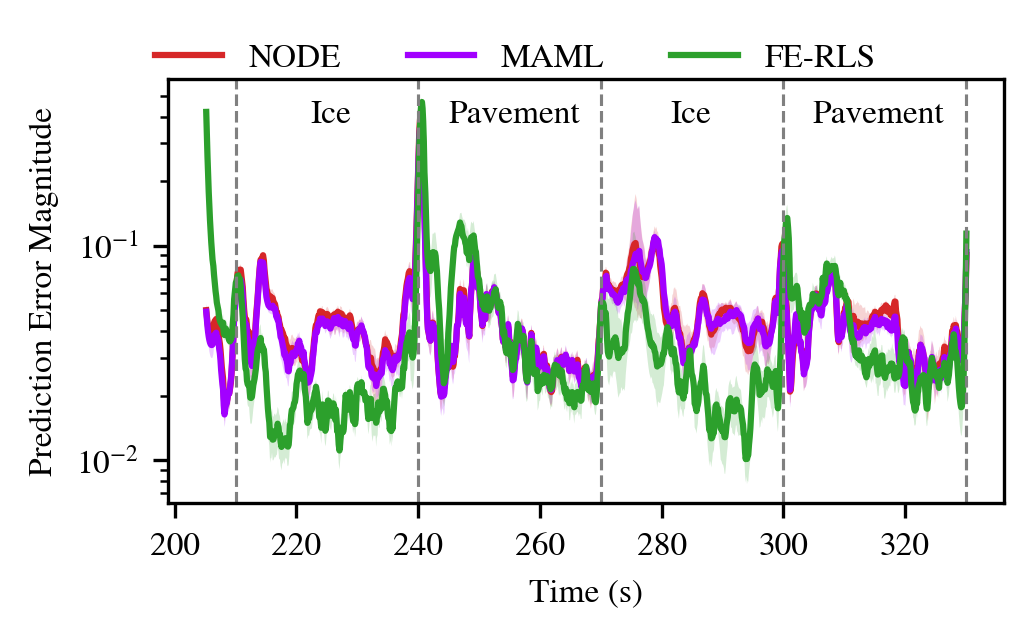}
    \vspace{-0.5cm}
    \caption{FE-RLS adapts to abrupt changes in terrain dynamics, outperforming baseline MAML and NODE. We compute the Euclidean norm of the error in predicted changes in state over time. }
    \label{fig: sim terrain adap}
\end{figure}

\begin{figure}
    \centering
    \includegraphics[keepaspectratio,width=\columnwidth]{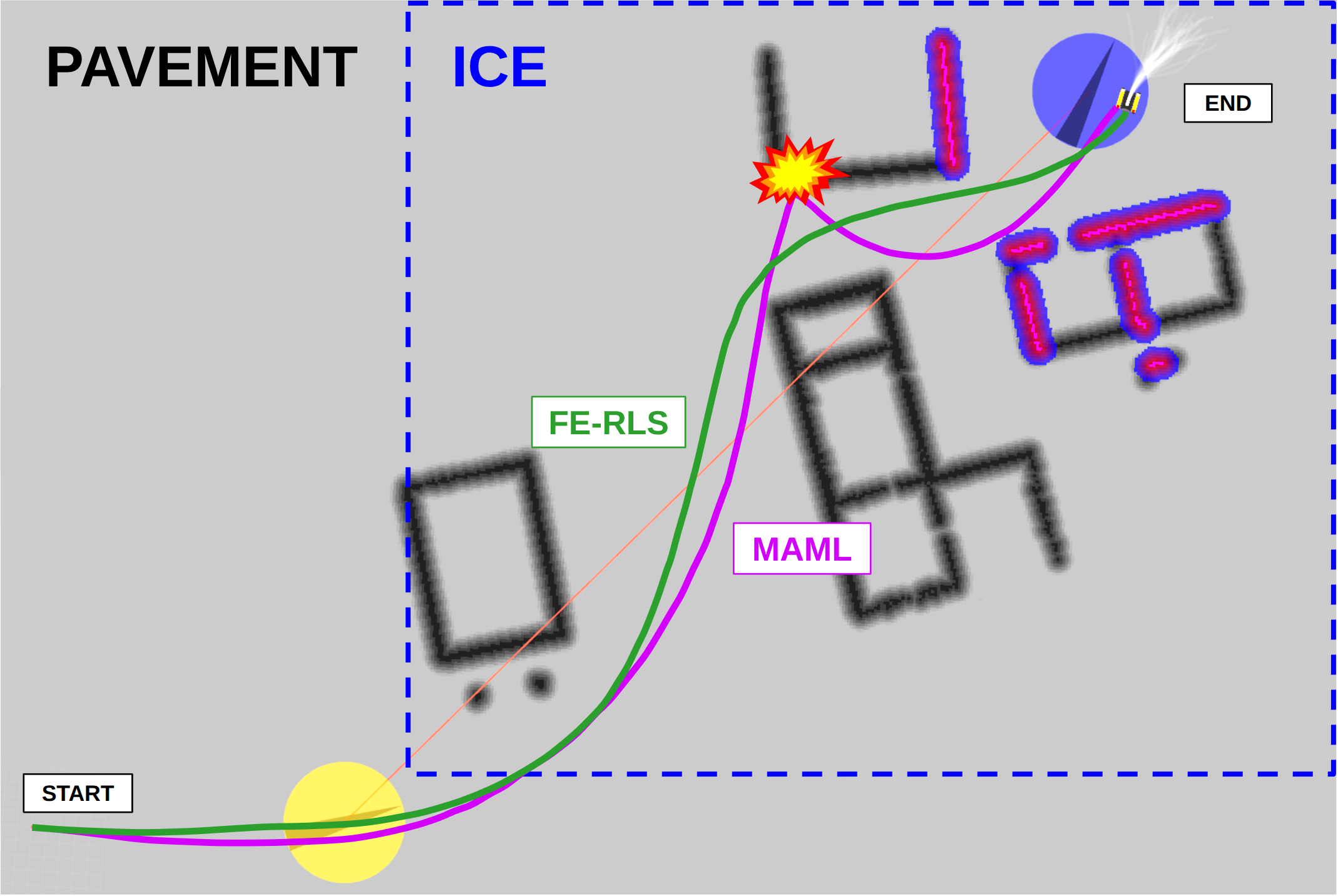}
    \vspace{-0.5cm}
    \caption{FE-RLS adapts to abrupt changes in terrain, quickly identifying the new dynamics, avoiding obstacles, and completing the mission. MAML fails to adapt, and collides with obstacles due to inaccurate MPPI rollouts.
    }
    \label{fig: simulator qual}
\end{figure}

% Experiment scope
We next evaluate adaptation in a high-fidelity Unity-based ROS simulator for unstructured navigation under controlled terrain changes.
This simulator provides a simulated Clearpath Warthog robot equipped with a 3D LiDAR, IMU, and camera.
The simulated robot runs a ROS based autonomy stack utilizing ground truth odometry, simultaneous localization and mapping based on OmniMapper~\cite{trevor2014omnimapper}, and waypoint navigation using MPPI~\cite{williams2018mppi}.
During evaluation, we plug and play a learned dynamics model for MPPI with one of our trained models.

Typically, the simulator enforces static terrain properties for an entire run.
To simulate abrupt terrain transitions (such as pavement to ice) and evaluate online adaptation, we modified the simulator to allow mid-trajectory changes in surface friction parameters.

% Dataset setup
We define two extreme parameter sets: one representing high-friction pavement and the other representing low-friction ice.
We then generate six intermediate terrains using convex combinations of these extremes, resulting in eight total terrains.
Each terrain corresponds to a different latent world state $w$, as in \eqref{eqn: system dynamics}.
For each terrain, we collect driving data using the simulator’s integrated MPPI controller guided by manually specified waypoints.
We collect $15$ minutes of data on each scene.
Each scene contains odometry (position, orientation, velocity, and angular rates) and commanded forward and angular velocities.
We transform inertial coordinates into the body frame to improve data efficiency by reducing trajectory sparsity.
The resulting dataset $D^{w_i}$ consists of input–output pairs $(x_t, v_t, \Delta t, x_{t+1} - x_{t})$ for training the learned models.

% Training
We train the function encoder, neural ODE, and MAML baselines on the collected datasets and use identical preprocessing pipelines. 
Training was performed on a workstation with an NVIDIA RTX 4090 GPU and Intel Core i9-13900K CPU.
The FE model used $k=8$ basis functions, with coefficients computed via least squares during training.
We train the MAML model using 5 inner steps with an inner learning rate of $10^{-2}$. 
See \cite{ingebrand2024FEnODEs} for ablations of the FE hyperparameters. 
Training required approximately 12 minutes for the FE, 4 minutes for the neural ODE, and 75 minutes for MAML.
For inference, all models used an RK4 numerical integrator.
In all experiments, the MPPI controller generated 3000 rollouts per iteration over a 10-second horizon with 0.1 second time steps. The mean inference times for FE, NODE, and MAML, evaluated over 10 trials, are $2.5 \times10^{-3}$, $3.2 \times 10^{-4}$, and $4.3  \times 10^{-4}$ seconds, respectively. 

% Quantitative - tracking performance - setup
Figure \ref{fig: sim k-step} evaluates all models trained using 10 random seeds on offline data from a simulated icy terrain. 
At each time, FE-RLS and MAML receive one input-output pair from the data, which each model uses to update its representation of the unknown dynamics.
Then, all models integrate the current state forward 15 steps into the future using historical control inputs from the dataset to simulate an MPPI rollout.
At each time step, we accumulate the prediction errors over the rollout.
We repeat this process starting from 30 random starting points in the dataset.
Then, we plot the median across each batch and random seed. The shaded region shows the 10th to 90th percentiles.

% Quantitative - tracking performance - explanation of results
Figure \ref{fig: sim k-step} highlights the predictive performance of our approach on an unseen simulated icy terrain.
Starting from zero prior knowledge, FE-RLS adapts within only a few seconds ($\approx 5$s), reaching prediction errors comparable to the pre-tuned batch function encoder but operating fully online. 
In contrast, the static neural ODE does not specialize to low-friction dynamics and accumulates large errors, while MAML adapts too slowly due to its reliance on gradient-based updates.

% Quantitative - 
The quantitative results in Figure \ref{fig: sim terrain adap} show that FE-RLS adapts to terrain changes between ice and pavement within seconds.
The static neural ODE does not specialize to the low-friction ice terrain, while MAML adapts slowly, requiring more samples for gradient-based updates than are available. Similar to the results shown in Figure \ref{fig: van der pol comparison}, the MAML signal overlaps the NODE signal throughout the experiment, indicating minimal adaptation. 
In scenarios with abrupt terrain shifts (pavement to ice), FE-RLS rapidly updates its coefficients to track the new dynamics, generating accurate rollouts, while MAML lags and produces rollouts inconsistent with the true system.
These results show that RLS enables function encoders to adapt on the same timescale as control, a property critical for real-time planning under rapidly varying terrain.

Figure~\ref{fig: simulator qual} shows an example scenario from our experiments that highlights the differences between FE-RLS and MAML performance.
While navigating an obstacle-rich urban environment, FE-RLS enables MPPI to maintain stable trajectories and complete waypoint navigation tasks under terrain transitions from high-friction pavement to low-friction ice.
In contrast, MAML shows delayed adaptation that leads to inaccurate rollouts and collisions.

\subsection{Hardware Experiments}

\begin{figure}
    \centering
    \includegraphics[keepaspectratio,width=\columnwidth]{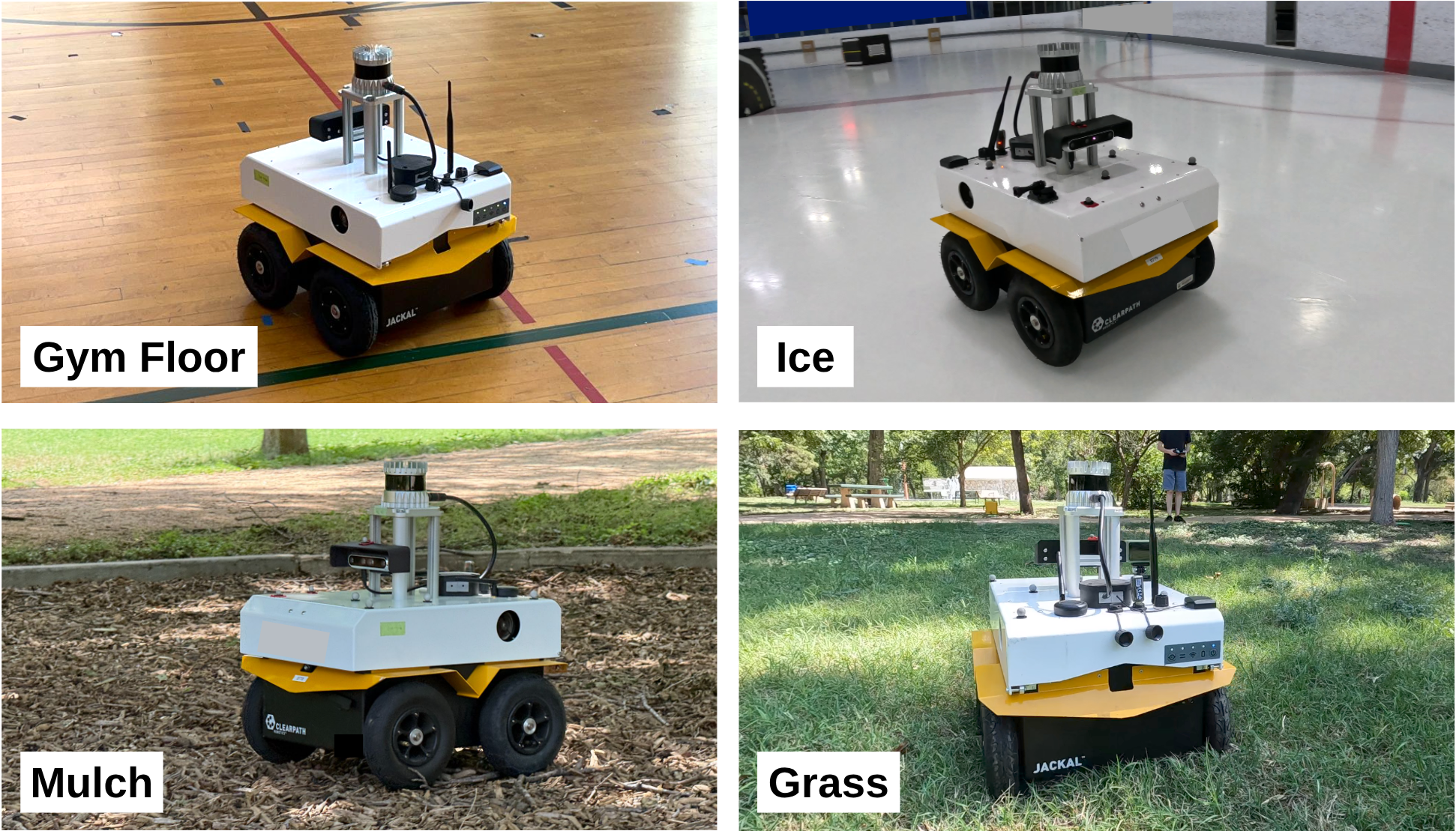}
    \vspace{-0.5cm}
    \caption{Representative examples of terrains in our dataset. We collect trajectory data across diverse indoor and outdoor terrains, including gym floor, mulch, grass, and a challenging terrain at a local ice rink.}
    \label{fig: hardware data collection}
\end{figure}

\begin{figure}
    \centering
    \includegraphics[keepaspectratio,width=\columnwidth]{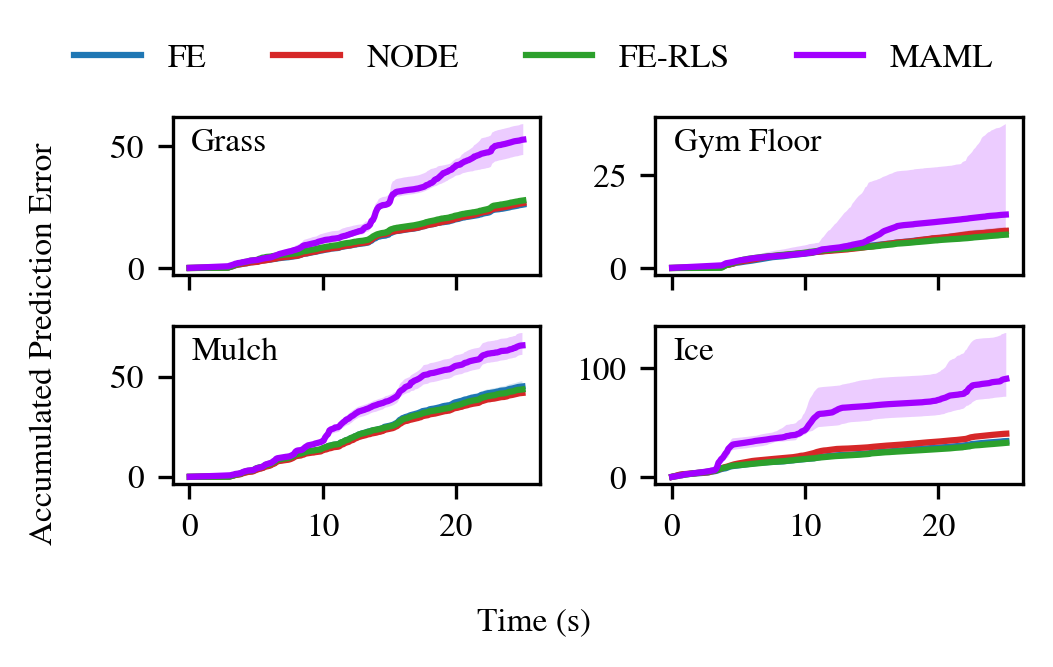}
    \vspace{-0.5cm}
    \caption{We evaluate model predictions on real-world trajectories from four representative indoor and outdoor terrains. Over time, MAML accumulates larger prediction error and variance due to unstable updates. We plot the median across models train using 10 random seeds. The shaded region shows the 10th to 90th percentiles.}
    \label{fig: hardware accum}
\end{figure}

\begin{figure}
    \centering
    \includegraphics[keepaspectratio,width=\columnwidth]{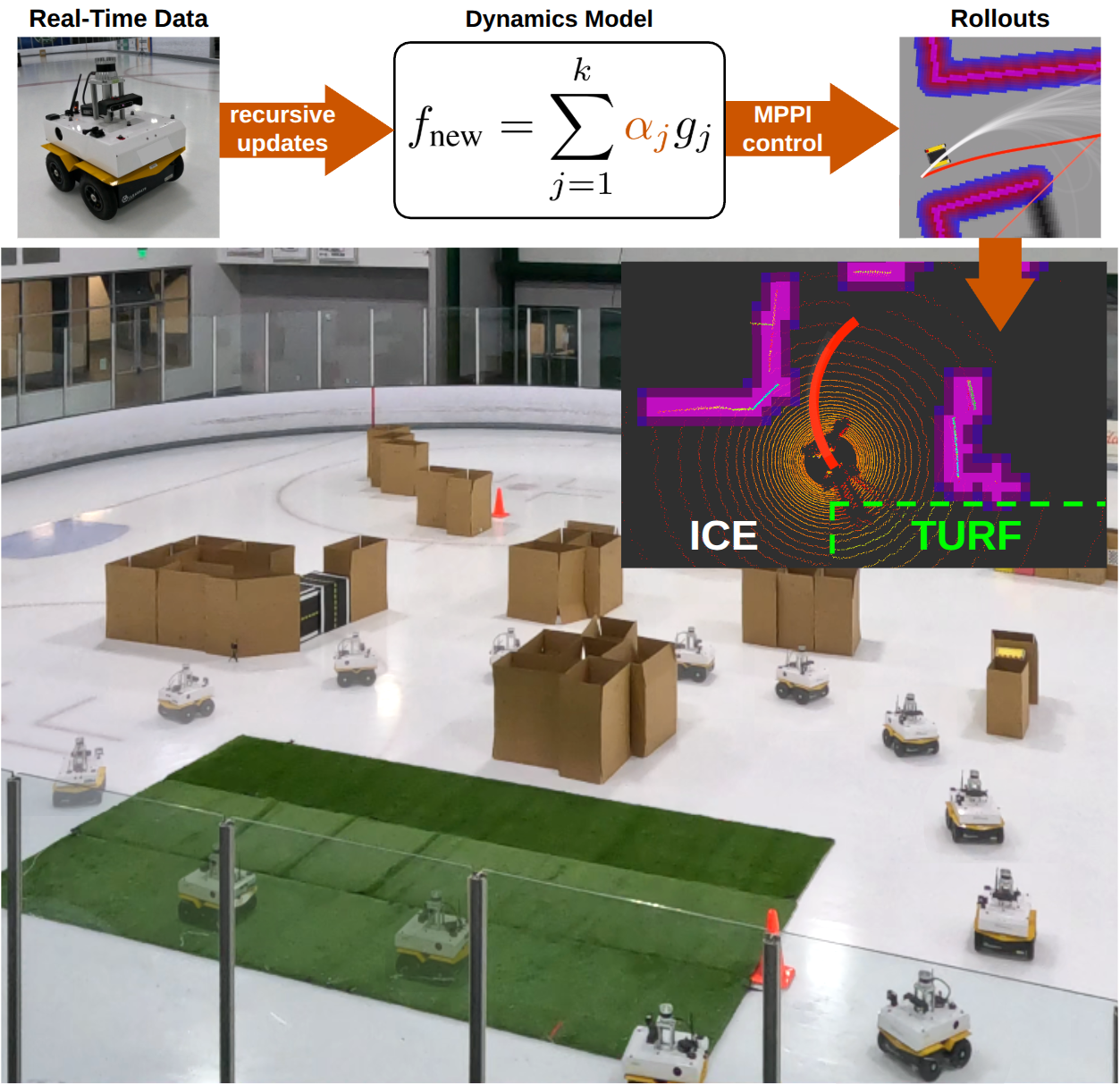}
    \vspace{-0.5cm}
    \caption{We demonstrate adaptation on a challenging mixed-terrain environment at a local ice rink. Real-time recursive updates of the FE-RLS model coefficients allow MPPI control to adapt seamlessly to abrupt terrain transitions during navigation.}
    \label{fig: hardware experiments}
\end{figure}

\begin{figure}
    \centering
    \includegraphics[keepaspectratio,width=\columnwidth]{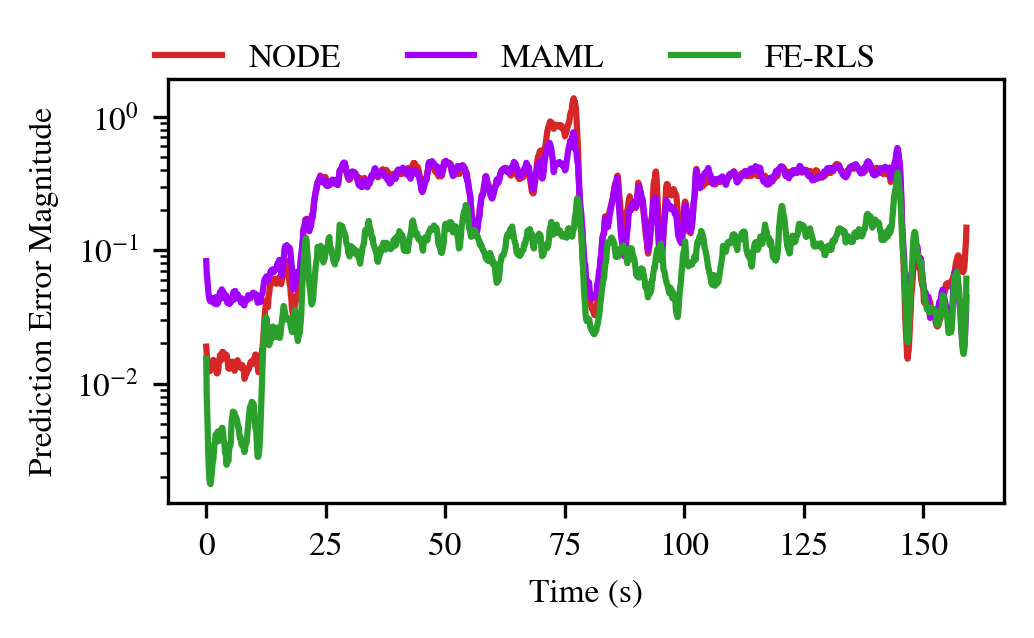}
    \vspace{-0.5cm}
    \caption{As the terrain switches between turf and ice, FE-RLS maintains lower prediction error than NODE and MAML. We evaluate each model on a trajectory recorded during real-time autonomy experiments from Figure \ref{fig: hardware experiments}.}
    \label{fig: hardware ice autonomy errors}
\end{figure}

% Data collection
Finally, we validate the method on a Clearpath Jackal robot in real outdoor environments.
These experiments evaluate feasibility on physical hardware and test performance under adversarial terrain conditions.
The Jackal performs all real-time computations on a Neousys Nuvo-7166GC computer with a NVIDIA T4 GPU. 
We collect odometry and command velocity data while remotely operating the Jackal across diverse indoor and outdoor terrains: grass, gym floor, ice, mulch, pavement, and turf (Figure~\ref{fig: hardware data collection}).
We estimate odometry measurements with DLIO~\cite{chen2022dlio} using an Ouster OS0-64u LiDAR and a Microstrain 3DM-GX5-AHRS IMU.
We record 15 minutes of data on grass, gym floor, mulch, and pavement, 16 minutes on turf, and 19 minutes on ice.
We process data from each terrain as in \ref{sec: sim results}, resulting in a dataset $D^{w_i}$ for training the learned dynamics models.

We train FE, FE-RLS, NODE, and MAML on the real-world datasets.
For offline evaluation, we train the FE model using $k = 8$ basis functions to remain consistent with our previous baselines. However, to mitigate computational issues with MPPI control on the Jackal hardware, we train a light-weight FE with $k = 3$ basis functions to test our approach in the full autonomy stack.
Reducing the number of basis functions decreases the forward model's computation time.
This ensures that the FE produces real-time trajectory rollouts. 
We anticipate that an implementation in a lower-level compiled language would dramatically improve forward computation time. 
We train the MAML model using 5 inner steps and an inner learning rate of $10^{-2}$. During evaluation, the gradient updates exploded ($> 10^{6}$) using the full 5 inner steps. To remain stable, we used only 1 inner step during evaluation to mitigate the sensitivity in the gradient updates.
In all experiments, the MPPI controller generated 100 rollouts per iteration over a 2.5-second horizon with 0.1 second time steps. 

In Figure \ref{fig: hardware accum}, we evaluate models trained using 10 random seeds on real-world trajectories from four representative terrains. Over time, FE-RLS, NODE, and the static FE (pre-tuned to each terrain using batch data) maintain smaller prediction errors than MAML. While NODE, static FE, and MAML all start with prior knowledge of the terrain, FE-RLS adapts starting from zero knowledge.

In Figure \ref{fig: hardware experiments}, we evaluate our terrain adaptation algorithm in a real-time control scenario on the Clearpath Jackal at a local ice rink. We design an obstacle-rich environment similar to the simulated test scenario in Figure \ref{fig: simulator qual}, and we create a terrain transition by placing a large patch of turf on the ice. The Jackal navigates around cardboard box obstacles using the same MPPI autonomy stack referenced in \ref{sec: sim results}. By manually placing waypoint goals throughout the scene, we lead the robot in a loop around the obstacles and across both terrains.

Figure \ref{fig: hardware experiments} shows a timelapse of the robot's trajectory when using the FE-RLS model. We observe that the model's recursive updates enable MPPI to maintain control of the robot across terrain changes between turf and ice. In our experiments, we compare the performance of FE-RLS with that of NODE and MAML. Although offline analysis reveals that FE-RLS maintains lower error across the scene, MPPI accommodates for errors in these models and is still able to control the robot. 

In Figure \ref{fig: hardware ice autonomy errors}, we evaluate FE-RLS, NODE, and MAML on a recorded trajectory from our autonomy experiments where the robot navigated over a terrain that switched between turf and ice.
As in Figure \ref{fig: sim terrain adap}, FE-RLS and MAML receive one input-output pair from the data for updating its representation of the current dynamics.
At each time step, FE-RLS, NODE, and MAML predict the change in state, given the current state and the historical control input. 
As expected, we see that FE-RLS maintains lower error than MAML and NODE.

% Impact of real world results
These hardware results demonstrate the broader applicability of our proposed method to real world off-road navigation.
Our method is capable of on-the-fly, real-time adaptation of learned dynamics models while running on resource constrained hardware.
Such functionality enables mobile robots to safely navigate a diverse set of complex terrains without explicitly traversing all of them ahead of time.

%%%%%%%%%%%%%%%%%%%%%%%%%%%%%%%%%%%%%%%%%%%%%%%%%%
\section{Conclusion \& Future Work}

We presented a method for real-time adaptation of learned dynamics models that combines function encoders with RLS. 
By updating function encoder coefficients online, our approach adapts at the same rate as the controller and integrates directly with MPPI.
Experiments on the Van der Pol system, a high fidelity Unity simulator, and Clearpath Jackal hardware showed that our method adapts more quickly and effectively than gradient-based meta-learning methods such as MAML, and achieves accuracy comparable to an idealized batch baseline while operating fully online.

Our hardware results also reveal important limitations that suggest areas for future research.
First, our experiments involved robots operating at moderate speeds, where differences in terrain dynamics are less pronounced.
We expect the benefits of fast adaptation to grow with higher speeds.
Second, our Python implementation runs inference slower than MAML, which reduces the effective control rate.
These findings point toward two directions for future work: extending evaluation to higher-speed ground vehicles where real-time adaptation is most critical, and optimizing the computational efficiency of forward inference.

\section*{Acknowledgments}
We would like to thank Chaz Henry and the staff at Ice \& Field at The Crossover for hosting our hardware ice experiments. We also would like to thank our colleagues Alex Begara, Tyler Ingebrand, Michal Pavol Podolinsky, Rwik Rana, and Nathan Tsoi for assisting with our experiments. 

\bibliographystyle{IEEEtran}
\bibliography{bibliography}

@inproceedings{trevor2014omnimapper,
  title={Omnimapper: A modular multimodal mapping framework},
  author={Trevor, Alexander JB and Rogers, John G and Christensen, Henrik I},
  booktitle={International Conference on Robotics and Automation},
  pages={1983--1990},
  year={2014},
  organization={IEEE}
}

@inproceedings{byravan2017se3,
  author={Byravan, Arunkumar and Leeb, Felix and Meier, Franziska and Fox, Dieter},
  title={{SE3-Pose-Nets}: Structured Deep Dynamics Models for Visuomotor Control},
  year={2018},
  booktitle={International Conference on Robotics and Automation},
}

@article{chen2022dlio,
  title={Direct {LiDAR}-Inertial Odometry: Lightweight {LIO} with Continuous-Time Motion Correction},
  author={Chen, Kenny and Nemiroff, Ryan and Lopez, Brett T},
  journal={International Conference on Robotics and Automation},
  year={2023},
}

@article{dallas2020online,
  title={Online terrain estimation for autonomous vehicles on deformable terrains},
  author={Dallas, James and Jain, Kshitij and Dong, Zheng and Sapronov, Leonid and Cole, Michael P and Jayakumar, Paramsothy and Ersal, Tulga},
  journal={Journal of Terramechanics},
  volume={91},
  pages={11--22},
  year={2020},
  publisher={Elsevier}
}

@inproceedings{djeumou2023drift,
  title={Autonomous drifting with 3 minutes of data via learned tire models},
  author={Djeumou, Franck and Goh, Jonathan YM and Topcu, Ufuk and Balachandran, Avinash},
  booktitle={International Conference on Robotics and Automation},
  year={2023},
}

@inproceedings{espinoza2019vehicle,
  title={Vehicle-Terrain Parameter Estimation for Small-Scale Unmanned Tracked Vehicles},
  author={Espinoza, Albert A and Torres-Filomeno, Jorge L and Monta{\~n}ez-S{\'a}nchez, Karla M and Ortiz-And{\'u}jar, {\'A}ngel J},
  booktitle={International Symposium on Measurement and Control in Robotics},
  year={2019},
}

@inproceedings{finn2017model,
  title={Model-agnostic meta-learning for fast adaptation of deep networks},
  author={Finn, Chelsea and Abbeel, Pieter and Levine, Sergey},
  booktitle={International Conference on Machine Learning},
  year={2017},
}

@inproceedings{gillespie2018learning,
  title={Learning nonlinear dynamic models of soft robots for model predictive control with neural networks},
  author={Gillespie, Morgan T and Best, Charles M and Townsend, Eric C and Wingate, David and Killpack, Marc D},
  booktitle={International Conference on Soft Robotics},
  year={2018},
}

@inproceedings{ingebrand2024FEnODEs,
 author = {Ingebrand, Tyler and Thorpe, Adam J. and Topcu, Ufuk},
 booktitle = {Advances in Neural Information Processing Systems},
 title = {Zero-Shot Transfer of Neural {ODE}s},
 year = {2024}
}

@inproceedings{ingebrand2024pmlr,
  title     = {Zero-Shot Reinforcement Learning via Function Encoders},
  author    = {Ingebrand, Tyler and Zhang, Amy and Topcu, Ufuk},
  booktitle = {International Conference on Machine Learning},
  year      = {2024},
}

@inproceedings{fe_transfer,
  author       = {Tyler Ingebrand and
                  Adam J. Thorpe and
                  Ufuk Topcu},
  title        = {Function Encoders: {A} Principled Approach to Transfer Learning in
                  Hilbert Spaces},
  booktitle      = {International Conference on Machine Learning},
  year         = {2025}
}

@inproceedings{kumar2021rma,
  author       = {Ashish Kumar and
                  Zipeng Fu and
                  Deepak Pathak and
                  Jitendra Malik},
  title        = {{RMA:} Rapid Motor Adaptation for Legged Robots},
  booktitle    = {Robotics: Science and Systems},
  year         = {2021},
}

@article{li2018multi,
  title={A multi-mode real-time terrain parameter estimation method for wheeled motion control of mobile robots},
  author={Li, Yuankai and Ding, Liang and Zheng, Zhizhong and Yang, Qizhi and Zhao, Xingang and Liu, Guangjun},
  journal={Mechanical Systems and Signal Processing},
  year={2018},
}

@inproceedings{liu2020learning,
  title={Learning on the job: Online lifelong and continual learning},
  author={Liu, Bing},
  booktitle={AAAI Conference on Artificial Intelligence},
  year={2020}
}

@article{liu2021lifelong,
  title={A lifelong learning approach to mobile robot navigation},
  author={Liu, Bo and Xiao, Xuesu and Stone, Peter},
  journal={Robotics and Automation Letters},
  year={2021},
}

@article{nie2022deep,
  title={Deep-neural-network-based modelling of longitudinal-lateral dynamics to predict the vehicle states for autonomous driving},
  author={Nie, Xiaobo and Min, Chuan and Pan, Yongjun and Li, Ke and Li, Zhixiong},
  journal={Sensors},
  year={2022},
}

@article{padmanabhan2018estimation,
  title={Estimation of terramechanics parameters of wheel-soil interaction model using particle filtering},
  author={Padmanabhan, Chandramouli and Gupta, Sayan and Mylswamy, Annadurai and others},
  journal={Journal of Terramechanics},
  year={2018},
}

@INPROCEEDINGS{pentzer2014online,
  author={Pentzer, Jesse and Brennan, Sean and Reichard, Karl},
  booktitle={American Control Conference}, 
  title={On-line estimation of vehicle motion and power model parameters for skid-steer robot energy use prediction}, 
  year={2014},
}

@article{reina2020terrain,
  title={Terrain estimation via vehicle vibration measurement and cubature {Kalman} filtering},
  author={Reina, Giulio and Leanza, Antonio and Messina, Arcangelo},
  journal={Journal of Vibration and Control},
  year={2020},
}

@article{spielberg2019neural,
  title={Neural network vehicle models for high-performance automated driving},
  author={Spielberg, Nathan A and Brown, Matthew and Kapania, Nitin R and Kegelman, John C and Gerdes, J Christian},
  journal={Science robotics},
  year={2019},
}

@INPROCEEDINGS{10885815,
  author={Ortiz, Kendric and DiPirro, Rachel and Thorpe, Adam J. and Oishi, Meeko},
  booktitle={Conference on Decision and Control}, 
  title={Online Learning of Dynamical Systems Using Low-Rank Updates to Physics-Informed Kernel Distribution Embeddings}, 
  year={2024},
}

@inproceedings{Wang2005GP,
 author = {Wang, Jack and Hertzmann, Aaron and Fleet, David J},
 booktitle = {Advances in Neural Information Processing Systems},
 title = {Gaussian Process Dynamical Models},
 year = {2005}
}

@inproceedings{ward2025FEMPPI,
  title={Online Adaptation of Terrain-Aware Dynamics for Planning in Unstructured Environments},
  author={William Ward and Sarah Etter and Tyler Ingebrand and Christian Ellis and Adam Thorpe and Ufuk Topcu},
  booktitle={Robotics: Science and Systems Workshop on Resilient Off-road Autonomous Robotics},
  year={2025},
}

@inproceedings{williams2017information,
  title        = {Information theoretic {MPC} for model-based reinforcement learning},
  author       = {Williams, Grady and Wagener, Nolan and Goldfain, Brian and Drews, Paul and Rehg, James and Boots, Byron and Theodorou, Evangelos},
  booktitle    = {International Conference on Robotics and Automation},
  year         = {2017},
}

@ARTICLE{williams2018mppi,
  author={Williams, Grady and Drews, Paul and Goldfain, Brian and Rehg, James and Theodorou, Evangelos},
  journal={Transactions on Robotics}, 
  title={Information-Theoretic Model Predictive Control: Theory and Applications to Autonomous Driving}, 
  year={2018},
}

@inproceedings{xian2021hyperdynamics,
  author       = {Zhou Xian and
                  Shamit Lal and
                  Hsiao{-}Yu Tung and
                  Emmanouil Antonios Platanios and
                  Katerina Fragkiadaki},
  title        = {HyperDynamics: Meta-Learning Object and Agent Dynamics with Hypernetworks},
  booktitle    = {International Conference on Learning Representations},
  year         = {2021},
}

@InProceedings{pmlr-v229-djeumou23a,
  title = 	 {How to Learn and Generalize From Three Minutes of Data: Physics-Constrained and Uncertainty-Aware Neural Stochastic Differential Equations},
  author =       {Djeumou, Franck and Neary, Cyrus and Topcu, Ufuk},
  booktitle = 	 {Conference on Robot Learning},
  year = 	 {2023},
}

@article{chen2018neuralODEs,
  title={Neural ordinary differential equations},
  author={Chen, Ricky TQ and Rubanova, Yulia and Bettencourt, Jesse and Duvenaud, David K},
  journal={Advances in Neural Information Processing Systems},
  year={2018}
}

@INPROCEEDINGS{xiao2025anycar,
  author={Xiao, Wenli and Xue, Haoru and Tao, Tony and Kalaria, Dvij and Dolan, John M. and Shi, Guanya},
  booktitle={International Conference on Robotics and Automation}, 
  title={AnyCar to Anywhere: Learning Universal Dynamics Model for Agile and Adaptive Mobility}, 
  year={2025},
  pages={8819-8825},
}

@INPROCEEDINGS{wang2024payattn,
  author={Wang, Sean J. and Zhu, Honghao and Johnson, Aaron M.},
  booktitle={International Conference on Robotics and Automation}, 
  title={Pay Attention to How You Drive: Safe and Adaptive Model-Based Reinforcement Learning for Off-Road Driving}, 
  year={2024},
  pages={16954-16960},
}

@INPROCEEDINGS{Meng-RSS-23, 
    AUTHOR    = {Xiangyun Meng AND Nathan Hatch AND Alexander Lambert AND Anqi Li AND Nolan Wagener AND Matthew Schmittle AND JoonHo Lee AND Wentao Yuan AND Zoey Chen AND Sameul Deng AND Greg Okopal AND Dieter Fox AND Byron Boots AND Amirreza Shaban}, 
    TITLE     = {{TerrainNet: Visual Modeling of Complex Terrain for High-speed, Off-road Navigation}}, 
    BOOKTITLE = {Proceedings of Robotics: Science and Systems}, 
    YEAR      = {2023}, 
    ADDRESS   = {Daegu, Republic of Korea}, 
    MONTH     = {July}, 
    DOI       = {10.15607/RSS.2023.XIX.103} 
}

\end{document}